# An Integrated, Conditional Model of Information Extraction and Coreference with Application to Citation Matching

**Ben Wellner**[*,†], **Andrew McCallum**[*], **Fuchun Peng**[*], **Michael Hay**[*]

[*]University of Massachusetts Amherst
Amherst, MA 01003 USA
{wellner, mccallum, fuchun, mhay}@cs.umass.edu

[†]The MITRE Corporation
202 Burlington Road
Bedford, MA 01730 USA

## Abstract

Although information extraction and coreference resolution appear together in many applications, most current systems perform them as independent steps. This paper describes an approach to integrated inference for extraction and coreference based on conditionally-trained undirected graphical models. We discuss the advantages of conditional probability training, and of a coreference model structure based on graph partitioning. On a data set of research paper citations, we show significant reduction in error by using extraction uncertainty to improve coreference citation matching accuracy, and using coreference to improve the accuracy of the extracted fields.

## 1 Introduction

Although information extraction (IE) and data mining appear together in many applications, their interface in most current systems would be better described as loose serial juxtaposition than as tight integration. Information extraction populates slots in a database by identifying relevant subsequences of text, but is usually unaware of the emerging patterns and regularities in the database. Data mining begins from a populated database, and is often unaware of where the data came from, or their inherent uncertainties. The result is that the accuracy of both suffers, and significant mining of complex text sources is beyond reach.

To address this problem we have previously advocated (McCallum & Jensen, 2003) the use of joint probabilistic models that perform extraction and data mining in an integrated inference procedure—the evidence for an outcome being the result of simultaneously making inferences both "bottom up" from extraction, and "top down" from data mining. Thus (a) intermediate hypotheses from both extraction and data mining can be easily communicated between extraction and data mining in a closed loop system, (b) mutually-reinforcing evidence from multiple sources will have the opportunity to be properly marshaled, (c) and accuracy and confidence assessment should improve.

In particular, we advocate creating these joint models as conditional random fields (CRFs) (Lafferty et al., 2001) that have been configured to represent relational data by using parameter tying in repeated patterns based on the structure of the data—also known as relational Markov networks (Taskar et al., 2002). In natural language processing, conditionally-trained rather than generatively-trained models almost always perform better because they allow more freedom to include a large collection of arbitrary, overlapping and non-independent features of the input without the need to explicitly represent their dependencies, *e.g.*, (Lafferty et al., 2001; Carreras et al., 2002; Pinto et al., 2003). In relational modeling, undirected graphical models allow greater freedom to represent auto-correlation and other relations without concern for avoiding circularities (Taskar et al., 2002; Neville et al., 2004). Both these modeling choices are in contrast to other related work in using directed, generatively-trained probabilistic models for information extraction (Marthi et al., 2003).

This paper presents a model, inference and learning procedure for a preliminary case of this extraction and data mining integration—namely information extraction and coreference on research paper citations.[1] Extraction in this context consists of segmenting and labeling the various fields of a citation, including title, author, journal, year, etc. Coreference (also known as identity uncertainty, record linkage or object consol-idation) is a key problem in creating databases created from noisy data. For example, without properly resolving "Stuart Russell," "S. Russell," and "Stuart Russel" to the same entity in a database, relational connections will be missing, and subsequent data mining will not find the patterns it should.

---

[1]We currently avoid calling this work a *joint* model of extraction and coreference, because we have not yet "closed the loop" by repeatedly cycling between extraction and coreference inference.

Using a data set of citations from CiteSeer (Lawrence et al., 1999), we present experimental results indicating that the type of integration we advocate does indeed hold promise—we show that modeling uncertainty about extraction of citation fields can improve coreference (in the face of different field orderings, abbreviations and typographic errors), and that leveraging predicted coreference can improve the extraction accuracy of these fields. Measurements of best-case scenarios show that there is yet further gain available to be found through this integration. Certainly further gains are expected from experiments that close the loop between extraction and coreference, rather than the limited, separate bi-directional results provided here.

Building on earlier work in coreference that assumes perfect extraction (McCallum & Wellner, 2003), we cast coreference as a problem in graph partitioning based on Markov random fields (Boykov et al., 1999; Bansal et al., 2002). The graphical model has cliques for pairs of citations, the log-clique-potential of which is the edge weight in the graph to be partitioned. These edge weights may be positive or negative, and thus the optimal number of partitions (equivalent to number of cited papers in this case) falls out naturally from the optimization function of the max-flow-min-cut partitioning. Later in this paper we provide statistical correlation results indicating that the redundancy in this "fully-connected graph partitioning" approach to coreference is more robust than a graphical model in which a "prototype yields observations."

In the model introduced in this paper, the graphical model consists of three repeated sub-structures: (1) a linear-chain representing a finite-state segmenter for the sequence of words in each citation (2) a boolean variable in a clique between each pair of segmented citations, representing graph-partitioning-style citation coreference decisions, (3) a collection of attribute variables once for each paper entity (that is, one for each partition in the coreference graph), noting that the number of these entity sub-structures is determined at inference time. Thus, this model is a special case of Model 1 in McCallum and Wellner (2003).

Inference within the linear chain is performed exactly by dynamic programming; inference within the fully-connected coreference is performed approximately by a simple graph partitioning algorithm, and inference within the entity attributes is performed exactly by exhaustive search. Across the three sub-structures, approximate inference is accomplished by variants of iterated conditional modes (ICM) (Besag, 1986). More precisely, approximate inference in the entire model proceeds as follows: (1) for each citation, $N$ segmentations with highest probability ($N$-best lists) are found by a variant of Viterbi and provided to coreference; (2) coreference decisions are inferred by approximate graph partitioning, integrating out uncertainty about the sampled $N$ segmentations; (3) these coreference decisions are used to infer the attributes of each paper by searching over all combinations of values suggested by each citation's segmentations; and finally (4) inference of citation segmentations are revised to make themselves more compatible with their corresponding entity attributes.

Joint parameter estimation in this complex model is intractable, and thus, as in inference, we perform parameter estimation somewhat separately for each of the three sub-structures. In all cases, estimation is iterative, consisting of BFGS quasi-Newton steps on a maximum a posteriori conditional likelihood, with a zero-mean spherical-variance Gaussian prior on the parameters. The parameters of the linear-chain are set to maximize the conditional likelihood of the correct label sequence, in the traditional fashion for linear-chain CRFs. The parameters for the distance function in graph partitioning are set to maximize the product of independent conditional likelihoods for each pairwise coreference decision. The parameters for the entity attributes are set by pseudolikelihood to maximize the likelihood of correct placement of edges between highest-accuracy citation segmentations and their true entity attributes.

We present experimental results on the four sections of CiteSeer citation-matching data (Lawrence et al., 1999). Using our integrated model, both extraction and coreference show significant reductions in error—by 25-35% for coreference and by 6-14% for extraction. We also provide some encouraging best-case experiments showing substantial additional potential gain that may come from more integrated joint inference and creation of additional features that leverage the capabilities of conditional probability models.

## 2 Model

This paper presents a method for integrated information extraction and coreference based on conditionally-trained, undirected graphical models—also known as conditional random fields (Lafferty et al., 2001). The model predicts entities and their attributes conditioned on observed text.

The model contains three types of repeated sub-structures with tied parameters. These three sub-structures are responsible for (1) information extraction, in the form of segmentation and labeling of word sequences in order to find (the fields of) each *mention* of an entity, (2) coreference among the mentions to discover when two mentions are referring to the same underlying *entity*, (3) representing the *attributes* of each entity and the dependencies among those attribute values. The attributes of each entity correspond to the canonical values that could be entered into database record fields, and the dependencies allow the model to represent expectations about what combinations of attributes would be expected in the world.

In general, *conditional random fields* (CRFs) are undirected graphical models that encode a conditional probability distribution using a given set of features. CRFs are defined as follows. Let $\mathcal{G}$ be an undirected model over sets of random variables $\mathbf{y}$ and $\mathbf{x}$. If $C = \{\{\mathbf{y}_c, \mathbf{x}_c\}\}$ is the set of cliques in $\mathcal{G}$, then CRFs define the conditional probability of an output labeling, $\mathbf{y}$ given the observed variables, $\mathbf{x}$ as:

$$p_\theta(\mathbf{y}|\mathbf{x}) = \frac{1}{Z(\mathbf{x})} \prod_{c \in C} \Phi(\mathbf{y}_c, \mathbf{x}_c), \quad (1)$$

where $\Phi$ is a potential function and $Z(\mathbf{x}) = \sum_{\mathbf{y}} \prod_{c \in C} \Phi(\mathbf{y}_c, \mathbf{x}_c)$ is a normalization factor. We assume the potentials factorize according to a set of features $\{f_k\}$, which are given and fixed, so that $\Phi(\mathbf{y}_c, \mathbf{x}_c) = \exp(\sum_k \lambda_k f_k(\mathbf{y}_c, \mathbf{x}_c))$. The model parameters are a set of real-valued weights $\Lambda = \{\lambda_k\}$, one weight for each feature.

CRFs have shown recent success in a number of domains especially in sequence modeling for natural language tasks (Lafferty et al., 2001; Sha & Pereira, 2003; Pinto et al., 2003; McCallum & Li, 2003; Sutton et al., 2004), often outperforming their generative counterparts. Their strength lies primarily in their ability to accommodate multiple, overlapping, non-independent features. By training the model to maximize the conditional probability of the output labels *given* the input values, CRFs avoid having to generate the observed variables or model their dependencies.

### 2.1 A CRF for Citation Extraction and Coreference

We now describe in detail a CRF for the integrated task of extracting fields and performing coreference among research paper citations. Let $\mathbf{x} = \{\mathbf{x}_1, ...\mathbf{x}_K\}$ be a set of observed citations ("mentions"), where each $\mathbf{x}_i = (x_{i1}, x_{i2}, ...)$ is a sequence of words forming the text of the citation. Let $\mathbf{s} = \{\mathbf{s}_1, ...\mathbf{s}_K\}$ be the corresponding set of label sequences, each label sequence, $\mathbf{s}_i$ indicating the membership of a citation word to a field (such as author, title or year). As a convenience, we also define citation fields $\mathbf{c} = \{\mathbf{c}_1, ...\mathbf{c}_K\}$, where $\mathbf{c}_i$ is a collection of variables containing the complete string value for each of the various fields of $\mathbf{x}_i$, deterministically agglomerated from the label sequence $\mathbf{s}_i$. Let $\mathbf{y} = \{y_{1,2}, ..y_{i,j}, ...y_{K-1,K}\}, i < j$ be a set of boolean coreference variables indicating whether or not citation $\mathbf{x}_i$ is referring to the same paper as citation $\mathbf{x}_j$; (note $y$ here is more specific than in Eq 1). Finally, let $\mathbf{a} = \{\mathbf{a}_1, ...\mathbf{a}_M\}$ be the set of attributes of each paper ("entities"), where $M$ is the number of underlying research paper entities. Here, entity attributes are field values, such as title and year, but canonicalized from their noisy appearance in the multiple coreference citation mentions.

As described above, the model consists of three repeated sub-structures: (1) a linear-chain on the elements of the $\mathbf{s}_i$ sequences conditioned on the $\mathbf{x}_i$ sequences, for finite-state citation segmentation and labeling (information extraction); (2) a fully-connected graph on the $\mathbf{x}_i$'s, with the binary coreference decision on the mention pair $(\mathbf{x}_i, \mathbf{x}_j)$ indicated by $y_{ij}$; note also that formally the graphical model requires potentials over all triples $(y_i, y_j, y_k)$ in order to enforce transitivity, but these potentials never actually have to be instantiated in inference by graph partitioning; (3) potentials measuring the compatibility between each mention's segmentation $\mathbf{c}_i$ and the attributes of its corresponding entity $\mathbf{a}_k$. This model is closely related to a combination of Models 1 and 2 in McCallum and Wellner (2003), where further details and background may be found. Inference in the model aims to find the mode of $P(\mathbf{a}|\mathbf{x}) = \sum_{\mathbf{c},\mathbf{s},\mathbf{y}} P(\mathbf{a}, \mathbf{y}, \mathbf{c}, \mathbf{s}|\mathbf{x})$, where the summed term is defined in Equation 2 by a product of potentials on cliques of the graphical model. Figure 1 shows an example graphical model with two coreferent citations and a singleton.

$$P(\mathbf{a}, \mathbf{y}, \mathbf{c}, \mathbf{s}|\mathbf{x}) = \frac{1}{Z_\mathbf{x}} \left( \prod_{i=1, j>i, k>j}^{K} \phi_0(y_{ij}, y_{ik}, y_{jk}) \right)$$
$$\left( \prod_{i=1}^{M} \prod_{j=1}^{K} \phi_1(\mathbf{a}_i, \mathbf{c}_j) \prod_{k>j}^{K} \phi_2(\mathbf{c}_j, \mathbf{c}_k, y_{jk}) \right)$$
$$\left( \prod_{i=1}^{K} \prod_{t=1}^{|\mathbf{s}_i|} \phi_3(s_{i(t-1)}, s_{it}, \mathbf{c}_i, x_i) \right)$$
$$(2)$$

### 2.2 Inference

Exact inference in this model is clearly intractable. We have, however, some clearly defined sub-structures within the model, and there is considerable previous work on inference by structured approximations—that is, performing some form of inference separately in different substructures and then (iteratively) integrating these results (Saul & Jordan, 1996; Yedidia et al., 2000; Wiegerinck, 2000).

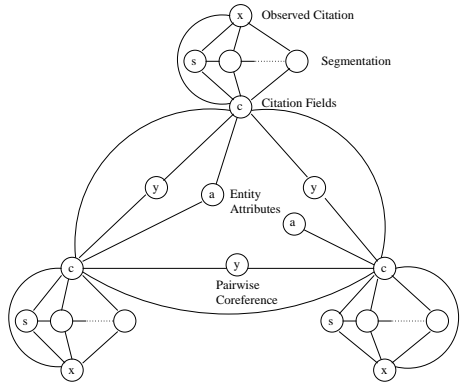

Figure 1: A model instance with three citations, two of which are co-referent.

Here we experiment with a particularly simple form of approximate inference: structured variants on iterated conditional modes (ICM) (Besag, 1986). In ICM, inference is performed by maximizing posterior probability of a variable, conditioned on all others, and repeatedly iterating over all variables. In its structured form, (possibly exact) inference may be performed on entire sub-structures of the model rather than a single variable, *e.g.* (Ying et al., 2002). In this paper we also use a variant we term *iterated conditional sampling*, in which, rather than selecting the single assignment of variables with maximum probability, several assignments are sampled (although not necessarily randomly) from the posterior and made available to subsequent inference. We expect that doing so makes the procedure less sensitive to local minima. These samples can also be understood as a strong compression of the exponentially-sized conditional probability tables that would have been sent as messages in structured belief propagation.

Inference in our model is performed as follows. First exact inference is performed independently for each label sequence $s_i$, conditioned on its corresponding citation word sequence $x_i$ using the Viterbi algorithm. Rather than selecting the single highest probability $s_i$ (mode), however, we find the $N$-best list of label sequences (sample). The citation fields $c_i$ are set deterministically from the sampled label sequences $s_i$.

Then coreference is accomplished by approximate inference via a greedy graph partitioning algorithm on the $y_{ij}$'s conditioned on the citation fields $c_i$, as described in McCallum and Wellner (2003), except that the edge weights in the graph are determined by "in-tegrating out the uncertainty in segmentation", that is, summing over all combinations of sampled label sequences, $s_i$ and their corresponding citation fields $c_i$. (Note that is is exactly where uncertainty in extraction is integrated into coreference.) Thus, the graph-partitioning edge weights between citations $c_i$ and $c_j$ are set to

$$w_{ij} = \log \left( \sum_{c_i, c_j, s_i, s_j} \phi_2(c_i, c_j, y_{ij}) \right.$$
$$\left. \left( \prod_{k=1}^{|s_i|} \phi_3(s_{i(t-1)}, s_{it}, c_i, x_i) \right) \left( \prod_{t=1}^{|s_j|} \phi_3(s_{j(t-1)}, s_{jt}, c_j, x_j) \right) \right) \quad (3)$$

Joint inference over all coreference decisions involves finding the mode of

$$P(y|x) = \frac{1}{Z_x} \left( \prod_{i=1, j>i, k>j}^{K} \phi_0(y_{ij}, y_{ik}, y_{jk}) \right) \left( \prod_{i,j>i}^{K} e^{w_{ij}} \right) \quad (4)$$

This problem is an instance of *correlation clustering*, which has sparked recent theoretical interest (Bansal et al., 2002; Demaine & Immorlica, 2003). Here we use a different approach to graph partitioning that bears more resemblance to agglomerative clustering — suitable for larger graphs with a high ratio of negative to positive edges. We search through this space of possible partitionings using a stochastic beam search. Specifically, we begin with each citation in its own cluster and select $k$ pairs of clusters to merge with probability proportional to the edge-weight between the two clusters. We examine each of these $k$ possible merges until one is found that results in an increase in the objective function. We repeat this until we reach a stage where none of the $k$ candidate merge pairs would result in an increased objective function value.

The final step to consider in our ICM-based inference is estimation of the attributes on entities, $a$, and a revisitation of citation segmentation given coreference and these entity attributes. Although our current experiments do not use entity attributes to affect coreference directly, they could do so by creating entity variables on the fly as coreference decisions are hypothesized. We seek to maximize

$$P(c, s, a|x) = \frac{1}{Z} \left( \prod_{i}^{M} \prod_{j}^{K} \phi_1(a_i, c_j) \right) \left( \prod_{j}^{K} \prod_{t=1}^{|s_j|} \phi_3(s_{j(t-1)}, s_{jt}, c_j, x_j) \right) \quad (5)$$

The segmentations $s$ and citation fields $c$ are selected only among the $N$-best segmentations, and the entity attributes are selected among the $N$-best citation fields of the coreferent citations for a given entity. Fortunately, there are few enough combinations that exact inference can be performed here. In our experiments, there are a maximum of 13 citation fields, (although typically much fewer), 2-21 citations in a coreferent cluster, and N was 5 or less. Given the attributes of an entity, we compute scores for all (entity, segmentation) pairs. The segmentation with highest score is selected as the best segmentation for the citation. These entity attributes are also scored by summing these highest scores for all citations. In the end, the entity attributes with the highest score are chosen as the canonical citation for this cluster and best segmentations are selected based on these attributes.

### 2.3 Parameter Estimation

Ideally we would perform parameter estimation by numerically climbing the gradient of the full, joint likelihood. This approach is not practical because complete inference in this model is intractable. In addition, our previous experience with coreference (McCallum & Wellner, 2003) indicates that learning parameters by maximizing a product of local marginals, $\prod_{i<j} P(y_{ij}|x_i, x_j)$, provides equal or superior accuracy to stochastic gradient ascent on an approximation of the full joint likelihood.

Following this success, here we train each substructure of the model separately, either as structured pseudo-likelihood, or simply independently. The parameters of the linear-chain CRF's potentials, $\phi_3(s_{i(t-1)}, s_{it}, \mathbf{c}_i\mathbf{x}_i)$, are set to maximize the joint probability of the correct label sequence, $P(\mathbf{s}_i|\mathbf{x}_i)$. We employ feature induction as part of this training (McCallum, 2003). The parameters of the coreference potentials on pairs of citations, $\phi_2(\mathbf{c}_i, \mathbf{c}_j, y_{ij})$, are set to maximize the product of local likelihoods of each pair, $\prod_{i<j} P(y_{ij}|x_i, x_j)$. The parameters of entity-attribute/citation potentials, $\phi_1(\mathbf{a}_i, \mathbf{c}_j)$, are set by pseudolikelihood to maximize the likelihood of correct placement of edges between citations and their true entity attributes. A spherical Gaussian prior with zero mean is used in all cases.

## 3 Experiments and Results

To evaluate our model, we apply it to a citation dataset from CiteSeer (Lawrence et al., 1999). The dataset contains approximately 1500 citations to 900 papers. The citations have been manually labeled for coreference and manually segmented into fields, such as author, title, etc. The dataset has four subsets of citations, each one centered around a topic (e.g. reinforcement learning). Within a section, many citations share common authors, publication venues, publishers, etc. The size of coreferent citation clusters has a skewed distribution; in three of the four subsets, at least 70% of the citations are singletons. The largest citation cluster consists of 21 citations to the same paper.

We present results on two different sets of experiments. First, we consider the coreference component of the model, which takes as input a sample of the N-best segmentations of each observation. We compare its performance to coreference in which we assume perfect labeling and in which we use no labeling at all. Second, we consider the segmentation performance of the model, which takes as input the citation clusters produced by the coreference component. We compare its accuracy to the baseline performance consisting of the top Viterbi segmentation from a linear-chain CRF.

### 3.1 Coreference Results

Good segmentation of author, title and other fields enables features that are naturally expected to be useful to accurate coreference. The pair-wise coreference potentials are a function of a wide range of rich, overlapping features. The features largely consider field-level similarity using a number of string and token-based comparison metrics.[2] Briefly, these metrics include various string edit distance measures, TFIDF over tokens, TFIDF over character n-grams as well hybrid methods that combine token TFIDF and string edit distance. We also used feature conjunctions (e.g. a feature that combines the author *and* title similarity measures). Some specialized features were developed for matching and normalizing author name fields as well as conference proceedings. Finally, we also included "global" features, based on string and token-based distance metrics, that looked at the entire citation. The features are a mix of real-valued and binary-valued functions.

We measure coreference performance at the pair-level and cluster-level. We report pairwise F1, which is the harmonic mean of pairwise precision and pairwise recall. Pairwise precision is the fraction of pairs in the same cluster that are coreferent; pairwise recall is the fraction of coreferent pairs that were placed in the same cluster. We also measure cluster recall, which is simply the ratio of the number of correct clusters to the number of true clusters. Note that cluster recall gives no credit for a cluster that is partially correct.

Table 1 summarizes coreference performance in terms of pair-wise F1 and cluster recall respectively. Results reported are on the indicated test section; the model was trained on the other three sections. We report the performance of our model using the $N$-best Viterbi segmentations for different values of $N$. Overall best results occur at $N = 9$. As a *lower bound* we also include the coreference performance when we include no segmentation information (*NoSeg*) and rely solely on the "global" features.

---

[2]We used the Secondstring package, some of the functions of which are described in (Cohen et al., 2003)

|         | Reinforce | Face  | Reason | Constraint |
|---------|-----------|-------|--------|------------|
| NoSeg   | 0.836     | 0.879 | 0.801  | 0.907      |
| N=1     | 0.972     | 0.974 | 0.946  | 0.961      |
| N=3     | 0.95      | 0.979 | 0.961  | 0.960      |
| N=7     | 0.948     | 0.979 | 0.951  | 0.971      |
| **N=9** | **0.982** | **0.967** | **0.960** | **0.971** |
| Labeled | 0.956     | 0.965 | 0.964  | 0.971      |
| Optimal | 0.995     | 0.992 | 0.994  | 0.988      |
| NoSeg   | 0.787     | 0.931 | 0.883  | 0.892      |
| N=1     | 0.913     | 0.971 | 0.920  | 0.931      |
| N=3     | 0.933     | 0.976 | 0.933  | 0.927      |
| N=7     | 0.933     | 0.976 | 0.937  | 0.950      |
| **N=9** | **0.947** | **0.969** | **0.937** | **0.951** |
| Labeled | 0.932     | 0.975 | 0.937  | 0.941      |
| Optimal | 0.98      | 0.996 | 0.993  | 0.976      |

Table 1: coreference performance measured by pair-wise F1 (upper part) cluster recall (lower part) using no segmentation (*NoSeg*), an average of the *N*-best Viterbi segmentations, and hand-labeled segmentations (*Labeled*). The *Optimal* result represents an upper-bound where the optimal pair-wise potential is chosen by an oracle.

Also included in the tables is the coreference performance when the hand-labeled segmentation is provided (*Labeled*). Note that the results using the $N = 9$ Viterbi segmentations are comparable to or higher than those using the correctly labeled segmentations—indicating that neither segmentation performance nor our technique for incorporating segmentation uncertainty are the inhibiting factor in improving coreference performance.

As an *upper-bound* experiment, we evaluate coreference performance assuming the model always chooses the optimal pair-wise potential from among the $N^2$ potentials. Thus, if the pair is coreferent, the potential is set to the maximum potential; if the pair is not coreferent, the potential is set to the minimum potential. The performance is near perfect (see *Optimal* in Tables 1).

Table 2 compares our best results to the results presented in (Pasula et al., 2003). As a baseline, we include the results of their implementation of the *Phrase matching* algorithm—a greedy agglomerative clustering algorithm where pair-wise citation similarity is based on the overlap in words and phrases (word bigrams). *RPM + MCMC* is their first-order, generatively-trained graphical model (see Related Work section). These results are not conclusive—better in some cases, worse in others—and larger more interesting datasets (more and larger citation clusters to leverage) may provide more interesting insights. Further leveraging conditional models' facility with feature engineering and induction may also prove helpful.

|              | Reinforce | Face  | Reason | Const. |
|--------------|-----------|-------|--------|--------|
| Phrase Match | 0.79      | 0.94  | 0.86   | 0.89   |
| RPM + MCMC   | 0.94      | 0.97  | 0.96   | 0.93   |
| CRF-Seg (N=9)| 0.947     | 0.969 | 0.937  | 0.951  |

Table 2: A comparison of cluster recall performance. The *Phrase Matching* and *RPM + MCMC* results are from (Pasula et al., 2003).

### 3.2 Segmentation Results

In leveraging coreference to improve extraction, we use a combination of local (e.g. word contains digits), layout, lexicon membership features (e.g. membership in a database of Bibtex records). See (Peng & McCallum, 2004) for a description of features. Segmentation performance is measured by the micro-averaged F1 across all fields, which approximates the accuracy of database fields. The segmentation component of the model was trained on a completely separate data set of citations (Peng & McCallum, 2004).

Table 3 shows the improved segmentation performance using coreference information. The results reported here only consider citations that were grouped together with at least one other citation (i.e. non-singletons), since these are the only citations whose segmentation we might hope to improve by using coreference. To test the significance of the improvements, we use McNemar's test on labeling disagreements (Gillick & Cox, 1989). Table 3 summarizes the significance test results. At the 95% confidence level (p-value smaller than 0.05), the improvements on the four datasets are significant.

|           | Reinforce | Face  | Reason | Constraint |
|-----------|-----------|-------|--------|------------|
| Baseline  | .943      | .908  | .929   | .934       |
| W/ Coref  | .949      | .914  | .935   | .943       |
| Err. Red. | .101      | .062  | .090   | .142       |
| P-value   | .0442     | .0014 | .0001  | .0001      |

Table 3: Comparison of segmentation performance on non-singleton citations using entity attributes generated through coreference vs. baseline segmentation.

We also explore the potential for improving segmentation performance by selecting among the $N$-best segmentations. For a given list of $N$ segmentations, we aim to select the segmentation closest to the true segmentation. The results in Table 4 show the optimal segmentation performance for different values of $N$.

We can see that there is further potential to improve segmentation based on optimally selecting segmentations.

|     | Reinforce | Face  | Reason | Constraint |
|-----|-----------|-------|--------|------------|
| N=1 | 0.936     | 0.911 | 0.912  | 0.933      |
| N=3 | 0.958     | 0.937 | 0.940  | 0.969      |
| N=5 | 0.962     | 0.948 | 0.946  | 0.975      |

Table 4: Optimal segmentation improvement for different values of N over all citations (including singletons).

## 4 Model Comparison

Our model consists of both explicit pair-wise coreference variables for each pair of citations, as well as explicit entity attribute variables for each group of co-referent citations. At inference time in our current experiments, however, coreference is driven solely by the pair-wise citation potentials. An alternative method would ignore the pair-wise potentials and consider only entity-citation potentials, creating entities at inference time as necessary.

If given a uniform prior over the number of entities, such a method would always chose to have a separate entity for each citation, and we must include a prior that prefers smaller numbers of entities (or equivalently, a penalty for generating each entity). Thus, the number of entities induced is a function of the "tension" between the entity-citation potentials (which depend on the observed citation strings, be highly parameterized, and be learned) and the prior (which do not). Intuitively, we wonder if this imbalance in expressiveness and learnability provides less delineative power and robustness than a more balanced alternative.

By contrast, a graph-partitioning, correlational-clustering approach to coreference includes potentials on citation pairs, and the number of entities emerges naturally from the tension between positive and negative edge weights in the graph—both of which can be conditionally trained and highly parameterized in their dependency on the observed citation strings. Also the compatibility between a set of hypothesized coreferent mentions is represented by a "mixture" rather than a single "prototype."

To explore these issues, we compare the two models using a randomly-generated sample of 100 partitionings of our citation dataset. Here we focus on the robustness of clustering coreferent citations; to remove the issues of tuning a prior over the number of entities, all randomly-generated partitions were constrained to have the correct number of clusters.

We then compute the objective function values of these partitionings according to both the pair-wise and entity-based models. The pair-wise model objection function is described above. The entity-based model's objective function is the product of the potentials between attributes of the entity and the field values of citations within a partition. The value of each entity attribute was generated by selecting the medoid field values from among the citations in the partition. (A medoid is the item in a cluster that has the minimum dissimilarity to all other items in the cluster—in this case, the similarity metric is a string-edit distance.) The potentials between citation-entity pairs were learned with exactly the same features and training procedure as in the pair-wise citation model.

For the randomly generated partitionings, we examine how well the probabilities for both models correlate ($r^2$ correlation) with the Pair F1 evaluation metric; see Table 5. Figure 2 shows scatter plots for the *Constraint* data set illustrating the correlation between the model objective function values and Pair F1.

We hypothesize that the pair-wise model correlates better partly due to the fact that the noise in edge potentials is ameliorated by averaging over $n$ potentials for each citation (one for each other citation) instead of only a single potential as is the case with the entity-based model. Perhaps even though the edge potentials in the entity-based model are expected to be less noisy (because the entity attributes are more "canonical"), averaging in the pair-wise model is still more robust. Further work is needed here to determine the effect of the quality of the generated entity attributes and learned potentials on the model's performance.

|        | Reinforce | Face  | Reason | Constraint |
|--------|-----------|-------|--------|------------|
| Pair   | 0.976     | 0.975 | 0.967  | 0.974      |
| Entity | 0.795     | 0.592 | 0.360  | 0.775      |

Table 5: Model objective function $r^2$ correlation with PairF1, in which the correct number of entities is provided.

## 5 Related Work

This paper is an example of integrating information extraction and data mining, as discussed in McCallum and Jensen (2003). Additional previous work in this area includes Nahm and Mooney (2000), in which data mining association rules are applied to imperfectly-extracted job posting data, and then used to improve the recall of subsequent information extraction.

Our work here is most related to the work of Pasula et al. (2003), who describe a first-order probabilistic model for citation segmentation and coreference. From the text of the citation, they extract the authors, title and publication type; (the rest of the citation is not explicitly segmented). They present the coreference performance of their model, and we reproduced those results in Table 2. In Marthi et al. (2003), their model is extended to include several more object classes (such as publication venues and publishers, however without experiments). In Milch et al. (2004), they further generalize by proposing a language for expressing first-order predicate logic for reasoning with unknown objects. Our work differs by several modeling choices: conditional instead of generative training, undirected instead of directed graphical models, identity uncertainty by graph partitioning (potentially with entity attributes also) instead of similarity to induced entities alone.

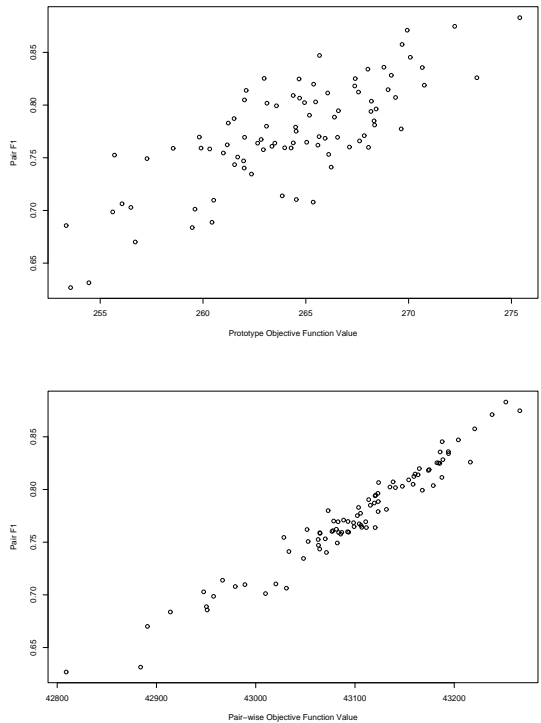

Figure 2: Constraint data set scatter plot for the entity model above and the pair-wise model below.

Our model can be understood as an instance of a relational Markov network Taskar et al. (2002), which have been employed in several complex relational domains and have demonstrated that considering relational dependencies between variables can improve performance. Loopy belief propagation has been used for inference at both learning and testing time in RMNs; it would be interesting to measure its suitability or lack thereof to the model in this paper.

Within information extraction, some work has been done with models that carry out multiple processing stages in a single model. Notably, Roth and tau Yih (2004) present a linear-programming approach to the task of simultaneously classifying entity types and their relations in text.

Our model here builds on the model of coreference described in McCallum and Wellner (2003), in which coreference is cast as a problem of graph partitioning. That paper presents a more complex approach to parameter estimation: maximum likelihood estimation over the full joint probability distribution. Here we have explored maximizing local likelihoods over pairs of citations as a more tractable method of parameter estimation. Here we have also described a model in which dependencies between entity attributes are modeled once-per-entity, rather than once-per-mention.

## 6 Conclusions

This paper recommends the integration of information extraction and data mining as a methodology that will enable the robust creation and mining of knowledge bases—doing so by allowing explicit reasoning about both noisy observations, uncertain coreference, and the underlying objects. Specifically here we advocate (1) conditional-probability training to allow free use of arbitrary non-independent features of the input, (2) undirected graphical models to represent autocorrelation and arbitrary possibly cyclic dependencies, (3) approximate inference and parameter estimation can be performed in these large graphical models by structured approximations. Furthermore, we have presented correlation results demonstrating the robustness of conditionally-trained graph partitioning.

Experimental results show significant improvements in coreference by using uncertainty information from extraction, and in extraction accuracy using results of coreference. Further gains are expected from current work in feature engineering that will take advantage of the flexibility allowed by conditional models, and with inference that "closes the loop" by repeatedly iterating between inference steps for extraction and coreference. Future plans include creating additional model structure for representing multiple entity types and the relations between them.

## Acknowledgments

This work was supported in part by the Center for Intelligent Information Retrieval and in part by The Central Intelligence Agency, the National Security Agency and National Science Foundation under NSF grant #IIS-0326249. Any opinions, findings and conclusions or recommendations expressed in this material are the author(s) and do not necessarily reflect those of the sponsor. We are also grateful to Charles Sutton and Brian Milch for helpful comments on a previous draft and David Jensen for helpful discussions.